\pdfoutput=1
\documentclass[11pt]{article}
\usepackage{amsmath} 
\usepackage{algorithm}
\usepackage{enumitem}
\usepackage{amssymb}
\usepackage{acl}
\usepackage{enumitem}
\usepackage{times}
\usepackage{latexsym}
\usepackage[absolute,overlay]{textpos}
\usepackage{booktabs} 
\usepackage{footmisc}
\usepackage{float}

\usepackage{inconsolata}

\usepackage{graphicx}

\usepackage{multirow}
\usepackage{array}
\usepackage{caption}
\usepackage{pifont}
\usepackage{amssymb}
\usepackage{hyperref}
\usepackage{graphicx}

\usepackage[T1]{fontenc} 
\usepackage[utf8]{inputenc} 
\usepackage[english]{babel} 
\usepackage[most]{tcolorbox}
\usepackage{appendix}
\usepackage{lipsum} 
\newtcolorbox{promptcontentbox}{
    colback=black!5!white,    
    colframe=black,           
    boxrule=0.5pt,            
    arc=0mm,                  
    outer arc=0mm,            
    breakable,                
    left=5pt,                 
    right=5pt,                
    top=5pt,                  
    bottom=5pt,               
    width=0.48\textwidth,         
    before skip=\medskipamount, 
    after skip=\medskipamount   
}

\title{Memory OS of AI Agent}

\author{  
  Jiazheng Kang \\
  Beijing University of Posts \\ and  Telecommunications \\
  \texttt{kjz@bupt.edu.cn}
  \And
  Mingming Ji \\
  Tencent AI Lab \\
  \texttt{matthhewj@tencent.com}
  \AND 
  Zhe Zhao \\
  Tencent AI Lab \\
  \texttt{nlpzhezhao@tencent.com}
  \And
  Ting Bai \thanks{ Corresponding author.}\\
  Beijing University of Posts \\ and Telecommunications \\
  \texttt{baiting@bupt.edu.cn}
}

\begin{document}
\maketitle

\begin{abstract}

Large Language Models (LLMs) face a crucial challenge from fixed context windows and inadequate memory management, leading to a severe shortage of long-term memory capabilities and limited personalization in the interactive experience with AI agents.
To overcome this challenge, we innovatively propose a \textbf{Memory} \textbf{O}perating \textbf{S}ystem, i.e., \textbf{MemoryOS},  to achieve comprehensive and efficient memory management for AI agents. Inspired by the memory management principles in operating systems, MemoryOS designs a hierarchical storage architecture and consists of four key modules: Memory Storage, Updating, Retrieval, and Generation. Specifically, the architecture comprises three levels of storage units: short-term memory, mid-term memory, and long-term personal memory. Key operations within MemoryOS include dynamic updates between storage units: short-term to mid-term updates follow a dialogue-chain-based FIFO principle, while mid-term to long-term updates use a segmented page organization strategy.
Our pioneering MemoryOS enables hierarchical memory integration and dynamic updating. Extensive experiments on the LoCoMo benchmark show an average improvement of 49.11\% on F1 and 46.18\% on BLEU-1 over the baselines on GPT-4o-mini, showing contextual coherence and personalized memory retention in long conversations. The implementation code is open-sourced at \url{https://github.com/BAI-LAB/MemoryOS}.


\end{abstract}

\section{Introduction}

Large Language Models (LLMs) demonstrate impressive capabilities in text comprehension and generation, but face inherent limitations in sustaining dialogue coherence due to their reliance on fixed-length contextual windows for memory management.
This fixed-length design inherently struggles to preserve continuity in dialogues with significant temporal gaps, often resulting in disjointed memory that manifests as factual inconsistencies and reduced personalization. 
Long-term memory coherence is critical in scenarios requiring persistent user adaptation, multi-session knowledge retention, or stable persona representation across extended interactions, where the limitations of fixed-length memory management in default LLMs become particularly acute, constituting a significant open challenge in the field.


To address this challenge, current memory mechanisms in default LLMs can be broadly categorized into three methodological types:
(1) \emph{Knowledge-organization} methods~\cite{xu2025mem,liu2023think}, such as A-Mem structure memory into interconnected semantic networks or notes to enable adaptive management and flexible retrieval; (2) \emph{Retrieval mechanism-oriented} approaches~\cite{emotional,zhong2024memorybank,li2024hello}, e.g., MemoryBank integrates semantic retrieval with a memory forgetting curve mechanism to allow long-term memory updating; and (3) \emph{Architecture-driven} methods~\cite{packer2023memgpt,chhikara2025mem0}, such as MemGPT use hierarchical structures with explicit read and write operations to dynamically manage context.
Although these diverse strategies typically operate in isolation, i.e., each focusing on single dimensions such as storage structure, retrieval machinism, or update strategies, no unified operating system has been proposed to enable systematic and comprehensive memory management for AI agents.

\begin{figure*}
    \centering
    \includegraphics[width=\textwidth=0.9]{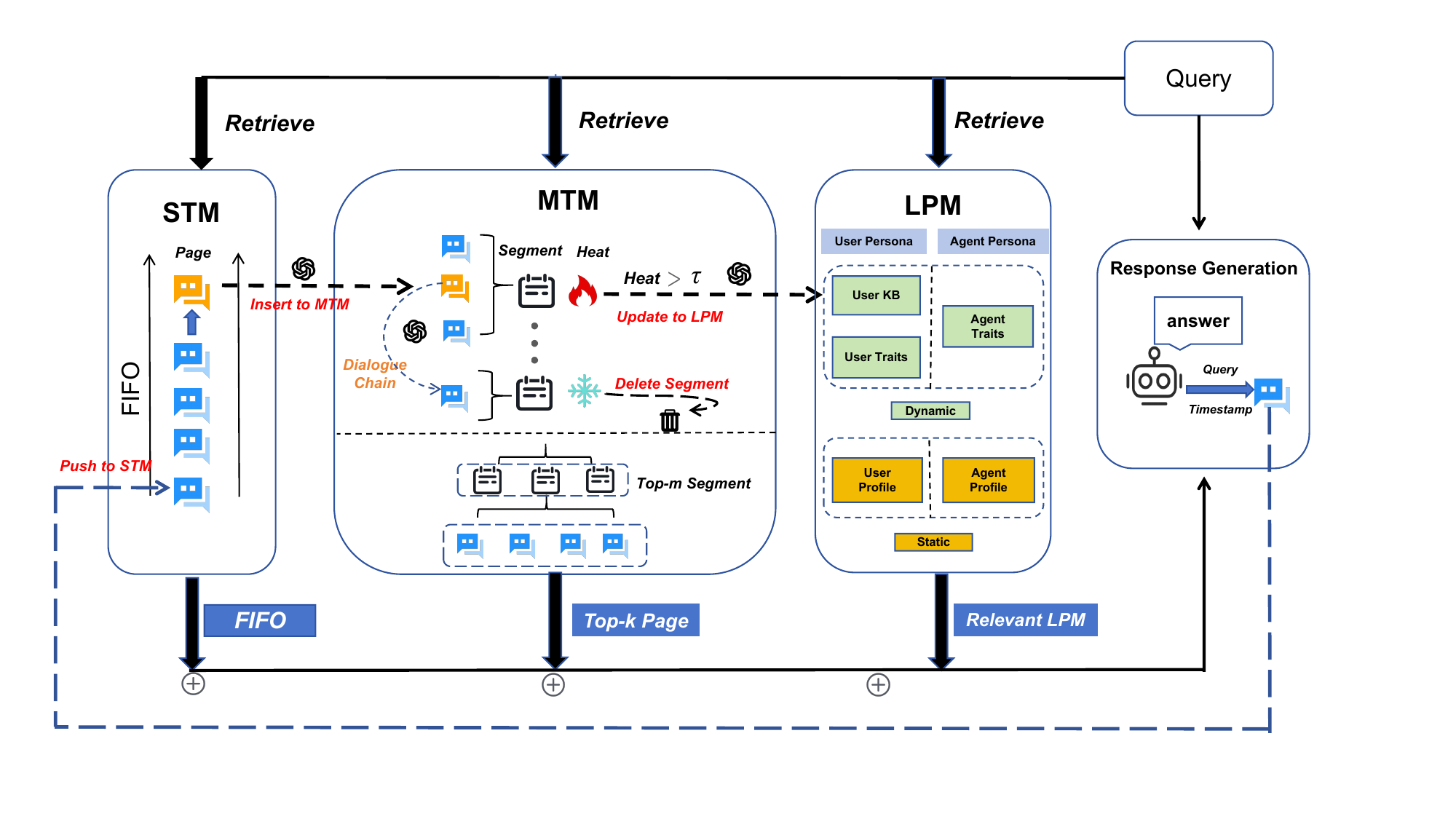} 
    \caption{The overview architecture of MemoryOS, including memory Store, Updating, Retrieval, Response.}
    \label{fig}
\end{figure*}

Inspired by memory management principles in operating systems, we pioneer the proposal of a comprehensive memory operating system, termed \textbf{MemoryOS}.
As illustrated in Fig.~\ref{fig}, MemoryOS comprises four core functional modules: memory \emph{Storage, Updating, Retrieval, and Generation}. 
Through their coordinated collaboration, the system establishes a unified memory management framework encompassing hierarchical storage, dynamic updating, adaptive retrieval, and contextual generation. 
Specifically, \emph{Memory Storage} organizes information into short-term, mid-term, and long-term storage units. \emph{Memory Updating} dynamically refreshes via a segmented paging architecture based on the dialogue-chain and heat-based mechanisms. \emph{Memory Retrieval} leverages semantic segmentation to query these tiers, then \emph{Response Generation} integrates retrieved memory information to generate coherent and personalized responses.
This synergistic workflow ensures holistic management of long-term conversational memory, enabling contextual coherence and personalized recall in extended dialogues.
The primary contributions of our work are summarized as:
\begin{itemize}
\item We make the first innovative attempt to introduce a systematic operating system, termed MemoryOS, for memory management, empowering AI agents with long-term conversational coherence and user persona persistence in long conversational interactions.

\item 
MemoryOS introduces a pioneering three-tier hierarchical memory storage architecture and integrates four core functional modules (i.e, storage, updating, retrieval, and generation) for memory management, enabling dynamic capture and evolution of user preferences across extended dialogues.

\item 
Comprehensive experiments validate MemoryOS's effectiveness and efficiency in maintaining response correctness and coherence across diverse benchmark datasets, demonstrating its capability to handle long conversational interactions.
\end{itemize}




\section{Related Work}
\subsection{Memory for LLM Agents}

Existing Large Language Models (LLMs) face fundamental challenges in handling complex scenarios requiring long-term coherence. These challenges stem from the inherent limitations of fixed-length designs, which struggle to maintain continuity in dialogues with significant temporal gaps, resulting in fragmented memory that manifests as factual inconsistencies and diminished personalization. 
Advancements in memory systems of LLMs addressing this problem can be broadly grouped into three categories: knowledge-organization, retrieval mechanism-oriented, and architecture-driven frameworks~\cite{zhang2024survey,wu2025human,du2025rethinking}.
\emph{Knowledge-organization} methods focus on capturing and structuring the intermediate reasoning states of large language models. For example, Think-in-Memory (TiM)~\cite{liu2023think} stores evolving chains-of-thought, enabling consistency through continual updates. A-Mem~\cite{xu2025mem} organizes knowledge into an interconnected note network that spans sessions. Grounded Memory~\cite{ocker2025ground} integrates vision-language models for perception, knowledge graphs for structured memory representation to enable context-aware reasoning in smart personal assistants.
\emph{Retrieval mechanism–oriented} approaches enrich the model with an external memory library. MemoryBank~\cite{zhong2024memorybank} logs conversations, events, and user traits in a vector database and refreshes them using a forgetting-curve schedule; AI-town~\cite{aitown2024} keeps memories in natural language and adds a reflection loop for relevance filtering. EmotionalRAG~\cite{emotional} retrieves memory entries by combining semantic similarity with the agent’s current emotional state using a hybrid strategy.
\emph{Architecture-driven} designs alter the core control flow to manage context explicitly. For example, MemGPT~\cite{packer2023memgpt} adopts an OS-like hierarchy with dedicated \texttt{read}/\texttt{write} calls, while Self-Controlled Memory (SCM)~\cite{wang2025scmenhancinglargelanguage} introduces dual buffers and a memory controller that gates selective recall.



\subsection{Memory Management in OS}

Modern operating systems (OS) use combined segment-page memory management to balance logical structure with efficient physical utilization.
Classic approaches like Multics ~\cite{bensoussan1972multics} organize memory into segments divided into pages, supporting efficient management, protection, and sharing. Segment metadata (size, access permissions) prevents external fragmentation ~\cite{bensoussan1972multics}, while paging reduces internal fragmentation ~\cite{denning1970virtual}. Advanced OS use priority-based eviction (e.g., LRU, working-set models) to maintain hot data ~\cite{denning1970virtual}, and Zheng et al. ~\cite{zheng2020segmentpage} show that combining coarse-grained segmentation with fine-grained paging minimizes overhead on many-core processors.

Inspired by the management in OS, our MemoryOS applies these principles by structuring its memory into logical segments (conversation topics) subdivided into pages. It uses heat-based prioritization to retain relevant content and efficiently discard or archive less-accessed information, enhancing context management and personalization.

\section{MemoryOS}
MemoryOS is a comprehensive memory management system for AI agents that dynamically updates memory and retrieves semantically relevant context, ensuring coherent and personalized interactions in long conversations.

\subsection{Overview Architecture}
The overview architecture of MemoryOS is illustrated in Fig.~\ref{fig}. It consists of four modules: memory storage, update, retrieval, and generation. 

\paragraph{Memory Storage:} This module is responsible for organizing and storing memory information by a three-tier hierarchical structure: Short-Term Memory (STM) for timely conversations, Mid-Term Memory (MTM) for recurring topic summaries, and Long-term Personal Memory (LPM) for user or agent preferences, ensuring memory integrity and effective utilization.
\paragraph{Memory Updating:} This module manages dynamic memory refreshing, including STM-to-MTM updates via dialogue-chain FIFO and MTM-to-LPM updates using a segmented page strategy with heat-based replacement.
\paragraph{Memory Retrieval:} This module retrieves relevant memory via specific queries, employing a two-tiered approach in MTM: semantic relevance identifies segments first, followed by retrieving pertinent dialogue pages. Finally, it combines persona attributes from LPM and contextual information from STM to generate responses, integrating all relevant memories for response generation.
\paragraph{Response Generation:} It processes the data and generates appropriate responses. It integrates retrieval outcomes from STM, MTM, and LPM into a coherent prompt, enabling the generation of contextually coherent and personalized responses.

\subsection{Memory Storage Module}
Memory storage module is implemented via a hierarchical structure consisting of three type store units, i.e., Short-Term Memory (STM), Mid-Term Memory (MTM), and Long-term Personal Memory (LPM) store units. 



\paragraph{Short-Term Memory (STM):} 
It stores real-time conversation data in units called \emph{dialogue pages}. Each dialogue page contains the user's query $Q$, the model's response $R$, and the timestamp $T$, structured as $page_i = \{Q_i, R_i, T_i\}$.
To ensure contextual coherence, a dialogue chain is constructed for each page to maintain contextual information in short-term continuous dialogue exchanges and ensure consistent context tracking. A \textbf{dialogue page} is defined as: 
\begin{equation}
page^{chain}_i= \{Q_i, R_i, T_i, meta^{chain}_i\},
\end{equation}
where the meta information 
is generated by an LLM in two steps: first, evaluating a new page’s contextual relevance to prior pages to determine chain linkage or resetting to the current page if semantically discontinuous; second, summarizing all chain pages into $meta^{chain}_i$.

\paragraph{Mid-Term Memory (MTM):} Inspired by the memory management principles in Operating Systems, it adopts a \emph{Segmented Paging} storage architecture. 
Dialogue pages with the same topic are grouped into segments, each containing multiple pages for a unique topic. The \textbf{segment} in MTM is defined as:
\begin{equation}
\small
\label{eq:insert}
segment_i = \{page_i \mid \mathcal{F}_{\text{score}}(page_i,segment_i) > \theta \},
\end{equation}
where the content of a segment is summarized by a LLM based on the related dialogue pages.  $\mathcal{F}_{\text{score}}$ measures the similarity between the dialogue page and the segment based on both semantic and keyword similarities, defined as: 
    \begin{equation}
        \mathcal{F}_\text{score} = \cos(\mathbf{e}_s, \mathbf{e}_p) + \mathcal{F}_{Jacard}(K_s, K_p),
    \label{eq:sim}
    \end{equation}
where $\mathbf{e}_s$ and $\mathbf{e}_p$ denote the embedding vectors of a segment and the dialogue page, $K_s$ and $K_p$ are the keyword sets summarized by LLMs in the segment and page, respectively.  $\mathcal{F}_{Jacard}$  is the Jacard similarity, defined as $\mathcal{F}_{Jacard} = \frac{|K_s \cap K_p|}{|K_s \cup K_p|}$.

Pages with similarity scores to a segment exceeding the threshold $\theta$ are merged into the same segment, ensuring topic coherence and semantic consistency within the segment.

\paragraph{Long-term Persona Memory (LPM):} 
This module ensures that both the user and the assistant maintain a persistent memory of important personal details and characteristics, ensuring consistency and personalization over long-term interactions.
It consists of two components: the User Persona and the AI Agent Persona. 
\begin{itemize}
\item \textbf{User Persona}.
The \textbf{User Profile} comprises a static component with fixed attributes (gender, name, birth year), a User Knowledge Base (\textbf{User KB}) that dynamically stores factual information extracted and incrementally updated from past interactions, and \textbf{User Traits} that contains the evolving interests, habits, and preferences of users over time.

\item \textbf{Agent Persona}. It contains the \textbf{Agent Profile}, which includes fixed settings like the role the AI agent assistant plays or its character traits, providing a consistent self-description.
The \textbf{Agent Traits} are dynamic attributes that develop through interactions with the user, potentially including new settings added by the user or interaction history, e.g., recommended items, during conversations.
\end{itemize}
\subsection{Memory Update Module}
Key updating operations include updates within each unit itself and the update mechanism from STM to MTM, MTM to LPM store units.

\paragraph{STM-MTM Update:} 
STM stores information in the form of dialogue pages in a queue with fixed length. We employ a First-In-First-Out (FIFO) update strategy for information migration to the Mid-Term Memory (MTM). 
New dialogue page is appended to the queue's end. When the STM queue reaches its maximum capacity, the oldest dialogue page is transferred from the STM to the MTM according to the FIFO principle.

\paragraph{MTM-LPM Update:}
MTM updates involve two operations, i.e., segment deletion and segment-to-LPM updates, both based on the Heat score of segments, defined as:
\begin{equation}
        Heat = \alpha \cdot N_{\text{visit}} + \beta \cdot L_{\text{interaction}} + \gamma \cdot R_{\text{recency}},
       \label{eq:heat_formula}
\end{equation}
where coefficients $\alpha$, $\beta$, and $\gamma$ determine the relative importance of each factor. $N_{\text{visit}}$ is the number of times the segment has been retrieved, $L_{\text{interaction}}$ denotes the total number of dialogue pages within the segment, and $R_{\text{recency}}$ is the time decay coefficient represents the duration since the last retrieval time of the current segment, defined as:
$R_{\text{recency}} = \exp \left(-\frac{\Delta t}{\mu}\right)$,
where $\Delta t$ is the time elapsed since the last access, measured in seconds, and $\mu$ is a configurable time constant (i.e., 1e+7).

These three metrics, i.e., retrieval count ($N_{\text{visit}}$), total dialogue pages ($L_{\text{interaction}}$), and time decay coefficient ($R_{\text{recency}}$), collectively represent the frequent access, high engagement, and recent use as core indicators of segment heat.
When the length of segments exceeds the maximum capacity, segments with the lowest heat are evicted. This mechanism ensures that the content stored in MTM can retain topics with high engagement frequency in long user conversations, preserving detailed dialogue content under these topics through a segmented paging structure.   

\paragraph{LPM Update:}
Segments with heat exceeding a threshold $\tau$ (i.e., 5) are transferred to LPM. Segments and their dialogue pages update User Traits, User KB, and Agent Traits.
Following the user traits in~\cite{mayipersona}, we construct personalized User Traits with 90 dimensions across three categories: basic needs and personality, AI alignment dimensions, and content platform interest tags. 
Then, extract and update these dimensions from segments and dialogue pages to autonomously evolve traits by LLMs.
Meanwhile, factual information relevant to the user and agent assistant is extracted and recorded into the User KB and Agent Traits, respectively. 
Both the User KB and Assistant Traits maintain a fixed-size queue (i.e., 100), employing a First-In-First-Out (FIFO) strategy. 
After memory transition, the number of pages $L_{interaction}$ in Eq.~\ref{eq:heat_formula} is reset to zero, causing the heat score of the segment to decline. This ensures continuous persona evolution without redundancy.

\subsection{Memory Retrieval Module}
The Memory Retrieval Module retrieves information from three parts: STM for recent context, MTM using a two-stage retrieval (segment and page level), and LPM for personalized knowledge.
Given a query from the user, the memory retrieval module retrieves from the stored
memory, i.e., STM, MTM, and LPM,  to return the most relevant information to generate the responses, defined as:
\begin{equation}
\mathcal{F}_\text{Retrieval}(STM, MTM, LPM | Q),
\end{equation}
where $\mathcal{F}_\text{Retrieval}$ is the retrieval strategies applied in three memory store units.

\paragraph{STM retrieval:} All dialogue pages are retrieved as STM holds the most recent contextual memory for the current conversation. 
\paragraph{MTM retrieval:} Inspired by psychological memory recall mechanisms~\cite{yuan-etal-2024-generative}, a two-stage retrieval process is employed: first selecting segments via a matching score (defined in Eq.~\ref{eq:sim}) to select top-m candidate segments, then selecting the top-k most relevant dialogue pages within these segments based on semantic similarity.
After retrieval, the segment’s visit counter $N_{\text{visit}}$ and recency factor $R_{\text{recency}}$ are updated. 
\paragraph{LPM retrieval:} The User KB and Assistant Traits each retrieve the top-10 entries with the highest semantic relevance to the query vector as background knowledge. All information in the User Profile, Agent Profile, and User Traits is utilized, as they store user preference information, agent characteristic information, and user-specific trait information.

\subsection{Response Generation Module}

Given the user query, the final prompt is constructed by integrating the above three types of retrieved content from STM, MTM and LPM, along with the user's query, form the final prompt input for the LLM to generate the final response.
The incorporation of memory from recent dialogue (STM), relevant conversation pages (MTM), and persona information (LPM)  ensures responses remain contextually coherent with current interactions, draw on historical dialogue details and summaries for depth, and align with user and assistant identities, respectively,  enabling coherent, accurate, and personalized interaction experiences of AI agent systems.

  
  


\section{Experiments}

\subsection{Experimental Settings}

\paragraph{Datasets.} 
We conduct our experiments on GVD~\cite{zhong2024memorybank} and LoCoMo benchmark~\cite{maharana-etal-2024-evaluating} datasets. The GVD dataset consists of multi-turn dialogues simulated from interactions between 15 virtual users and an assistant over a 10-day period, covering at least two topics per day. 
The LoCoMo benchmark is specifically designed for assessing long-term conversational memory capabilities, consisting of ultra-long dialogues averaging 300 turns and about 9K tokens per conversation. Questions are categorized into four types: Single-hop, Multi-hop, Temporal, and Open-domain, to systematically evaluate the memory abilities of LLMs.

\paragraph{Evaluation Metrics.} 
For the GVD dataset, we use three evaluation metrics: Memory Retrieval Accuracy (Acc.), Response Correctness (Corr.), and Contextual Coherence (Cohe.).
Memory Retrieval Accuracy is evaluated as a binary indicator (0 or 1), while Correctness and Coherence are assessed on a three-point scale (0, 0.5, or 1). All evaluations on the GVD dataset are automatically scored by the DeepSeek-R1~\cite{deepseekai2025deepseekr1incentivizingreasoningcapability}. On the LoCoMo benchmark, standard F1 and BLEU-1~\cite{papineni2002bleu} are employed to evaluate the model’s performance.


\paragraph{Compared Methods.}
We compare MemoryOS with representative memory methods, including:

\paragraph{TiM (Think-in-Memory)~\cite{liu2023think}:}
This approach mimics human memory by storing reasoning outcomes instead of raw dialogues. It uses locality-sensitive hashing (LSH) to retrieve relevant context before generating responses and updates memory through post-hoc reflection. TiM manages memory via insertion, forgetting, and merging to reduce redundant reasoning and improve consistency.
\paragraph{MemoryBank~\cite{zhong2024memorybank}:} This framework dynamically adjusts memory strength based on the Ebbinghaus Forgetting Curve, prioritizing important content over time. It further builds a user portrait through continuous interaction analysis to support personalized responses.
\paragraph{MemGPT~\cite{packer2023memgpt}:} This method introduces a dual-tier memory, featuring a main context for fast access and an external context for long-term storage. This design aims to enable scalable memory extension beyond the fixed context window of LLMs.
\paragraph{A-Mem (Agentic Memory)~\cite{xu2025mem}:} it dynamically generates structured notes and links them to form interconnected knowledge networks, enabling continuous memory evolution and adaptive management for LLMs.
\paragraph{\textbf{MemoryOS}:} It is a comprehensive memory management framework. Through coordinated collaboration with four core functional modules: memory Storage, Updating, Retrieval, and Generation. MemoryOS achieves dialogue coherence and user persona persistence in long interactions.

\begin{table}[t]
\centering
\small
\setlength{\tabcolsep}{2.5pt}
\caption{Comparison results on the GVD dataset.}
\begin{tabular}{llccc}
\toprule
\textbf{Model} & \textbf{Method} & \textbf{Acc. $\uparrow$} & \textbf{Corr. $\uparrow$} & \textbf{Cohe.$\uparrow$}\\
\midrule
\multirow{5}{*}{GPT-4o-mini}
& TiM          & 84.5 & 78.8 & 90.8\\
& MemoryBank   & 78.4 & 73.3 & 91.2\\
& MemGPT       & 87.9 & 83.2 & 89.6\\
& A-Mem        & \underline{90.4} & \underline{86.5} & \underline{91.4}\\
& \textbf{Ours}& \textbf{93.3} & \textbf{91.2} & \textbf{92.3}\\
\midrule
\multicolumn{2}{c}{Improvement (\%)} &3.2$\%\uparrow$&5.4$\%\uparrow$&1.0$\%\uparrow$\\
\midrule
\multirow{5}{*}{Qwen2.5-7B}
& TiM          & 82.2 & 73.2 & 85.5\\
& MemoryBank   & 76.3 & 70.3 & 82.7\\
& MemGPT       & 85.1 & 80.2 & 86.9\\
& A-Mem        & \underline{87.2} & \underline{79.5} & \underline{87.8}\\
& \textbf{Ours}& \textbf{91.8} & \textbf{82.3} & \textbf{90.5}\\
\midrule
\multicolumn{2}{c}{Improvement (\%)} &5.3$\%\uparrow$&3.5$\%\uparrow$&3.1$\%\uparrow$\\
\bottomrule
\end{tabular}
\label{tab:gvd_results}
\end{table}

\begin{table*}
\centering
\scriptsize
\setlength{\tabcolsep}{2pt}
\caption{LoCoMo dataset comparison with per-category scores and average
ranks. A-Mem refers to the results reported in the original paper. A-Mem* represents our implementation results under the same experimental environment as our model.}
\begin{tabular}{ll|cc|cc|cc|cc|c|c}
\toprule
\multirow{2}*{\textbf{Model}} & \multirow{2}*{\textbf{Method}}
& \multicolumn{2}{c|}{Single Hop}
& \multicolumn{2}{c|}{Multi Hop}
& \multicolumn{2}{c|}{Temporal}
& \multicolumn{2}{c|}{Open Domain}
& \textbf{Avg.\ Rank $\downarrow$} & \textbf{Avg.\ Rank $\downarrow$}\\
& & F1 $\uparrow$ & BLEU-1 $\uparrow$ & F1 $\uparrow$& BLEU-1 $\uparrow$& F1 $\uparrow$& BLEU-1 $\uparrow$ & F1 $\uparrow$& BLEU-1 $\uparrow$
& (F1) & (BLEU-1)\\
\midrule
\multirow{6}{*}{GPT-4o-mini}
& TiM          & 16.25 & 13.12 & 18.43 & 17.35 &  8.35 &  7.32 & 23.74 & 22.05 & 3.8 & 4.0\\
& MemoryBank   &  5.00 &  4.77 &  9.68 &  6.99 &  5.56 &  5.94 &  6.61 &  5.16 & 5.0 & 5.0\\
& MemGPT       & \underline{26.65} & \underline{17.72} & 25.52 & 19.44 &  \underline{9.15} &  7.44 & \underline{41.04} & \underline{34.34} & 2.2 & 2.5\\
& A-Mem        & 27.02 & 20.09 & 45.85 & 36.67 & 12.14 & 12.00 & 44.65 & 37.06 & --  & -- \\
& A-Mem*  & 22.61 & 15.25 & \underline{33.23} & \underline{29.11} &  8.04 &  \underline{7.81} & 34.13 & 27.73 & 3.0 & 2.5\\
& \textbf{Ours}& \textbf{35.27} & \textbf{25.22} & \textbf{41.15} & \textbf{30.76} & \textbf{20.02} & \textbf{16.52} & \textbf{48.62} & \textbf{42.99} & \textbf{1.0} & \textbf{1.0}\\
\midrule
\multicolumn{2}{c|}{Improvement (\%)} & 32.35\%$\uparrow$ & 42.33\%$\uparrow$ & 23.83\%$\uparrow$ &  5.67\%$\uparrow$ & 118.80\%$\uparrow$ & 111.52\%$\uparrow$ & 18.47\%$\uparrow$ & 25.19\%$\uparrow$ & -- & -- \\
\midrule
\multirow{6}{*}{Qwen2.5-3B}
& TiM          &  4.37 &  5.01 &  2.54 &  3.21 &  6.20 &  5.37 &  6.35 &  7.34 & 4.3 & 3.5\\
& MemoryBank   &  3.60 &  3.39 &  1.72 &  1.97 &  6.63 &  6.58 &  4.11 &  3.32 & 4.8 & 4.8\\
& MemGPT       &  5.07 &  4.31 &  2.94 &  2.95 &  7.04 &  7.10 &  7.26 &  5.52 & 2.8 & 3.8\\
& A-Mem        & 12.57 &  9.01 & 27.59 & 25.07 &  7.12 &  7.28 & 17.23 & 13.12 & --  & -- \\
& A-Mem*       & \underline{10.31} &  \underline{8.76} & \underline{16.31} & \underline{11.07} &  \underline{6.94} &  \underline{7.31} & \underline{12.34} & \underline{10.62} & 2.3 & 2.0\\
& \textbf{Ours}& \textbf{23.26} & \textbf{15.39} & \textbf{21.44} & \textbf{14.95} & \textbf{10.18} &  \textbf{8.18} & \textbf{26.23} & \textbf{22.39} & \textbf{1.0} & \textbf{1.0}\\
\midrule
\multicolumn{2}{c|}{Improvement (\%)} &125.61$\%\uparrow$ & 75.68$\%\uparrow$ & 31.45$\%\uparrow$ & 35.05$\%\uparrow$ & 46.69$\%\uparrow$ & 11.90$\%\uparrow$ & 112.56$\%\uparrow$ & 110.83$\%\uparrow$ & -- & -- \\
\bottomrule
\end{tabular}
\label{tab:LoCoMo_results}
\end{table*}

\begin{table}
\centering
\small
\caption{Efficiency analysis on LoCoMo benchmark (quantified by LLM call and recalled tokens counts).}
\begin{tabular}{lccc}
\toprule
\textbf{Method} & \textbf{Tokens} & \textbf{Avg.Calls}& \textbf{Avg. F1} \\
\midrule
MemoryBank & 432 & 3.0&6.84 \\
TiM & 1,274 & 2.6 &18.01\\
MemGPT & 16,977 & 4.3&29.13 \\
A-Mem* & 2,712 & 13.0 &26.55\\
\textbf{Ours} & \textbf{3,874} & \textbf{4.9} &\textbf{36.23} \\
\bottomrule
\end{tabular}
\label{tab:consumption_analysis}
\end{table}

\begin{figure*}
    \centering
    \includegraphics[width=0.95\linewidth]{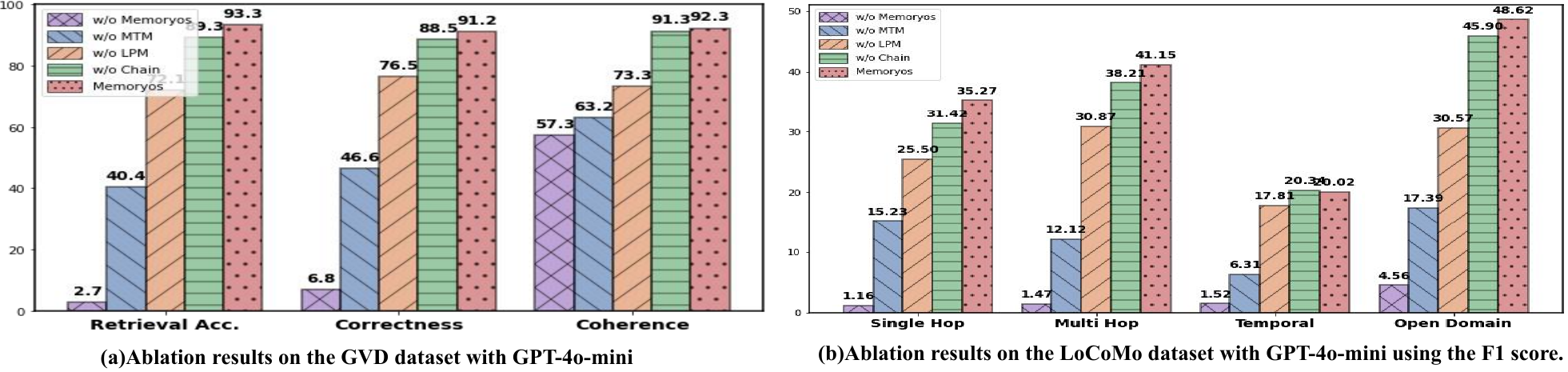}
    \caption{The ablation study on the GVD and LoCoMo benchmark datasets.}
    \label{fig:ablation_gvd}
\end{figure*}

\paragraph{Implementation Details.}
The experiments are conducted on hardware equipped with 8-H20 GPUs.
The fixed length of the dialogue page queue in STM is 7. 
The maximum length of segments in MTM is set to 200.
The maximum capacity for both the User KB and Agent Traits is set to 100 entries.
The predefined $Heat$ threshold $\tau$, which controls the information from the MTM to the LPM, is set to 5. 
The values of $\alpha$, $\beta$, and $\gamma$ in Eq.~\ref{eq:heat_formula} are equality set to 1. 
For memory retrieval, the number of retrieval top-$m$ segments was set to 5, and the hyperparameter top-$k$ for retrieved dialogue page was set to 5 and 10 on the GVD and LoCoMo datasets. 
The similarity value of $\theta$ in Eq.~\ref{eq:insert} is 0.6, and the time constant $\mu$ is 1e+7.  

\subsection{Main Results}
The experimental resutls in GVD and LoCoMo benchmark datasets are shown in Table~\ref{tab:gvd_results} and Table~\ref{tab:LoCoMo_results}. We have the following observations:



(1) Among all memory methods, MemoryBank performs the worst. This indicates that simply applying memory decay mechanisms is insufficient for managing conversational memory effectively. 
TiM outperforms MemoryBank by mitigating repetitive reasoning by saving ``thoughts" rather than raw turns, but its single-stage hash retrieval cannot preserve cross-topic dependencies. 

(2) A-Mem and MemGPT demonstrate relatively strong performance in long-form dialogue, But both of them lack systematic memory management mechanisms, giving rise to certain issues. For instance, MemGPT extends context via OS-style paging, yet its flat FIFO queue causes topic mixing as dialogue length grows; A-Mem organizes memories into a graph that enriches semantics, but the heavy, multi-step link generation inflates latency and error accumulation.
By contrast, our Memory OS fuses a hierarchical STM/MTM/LPM architecture via segmented paging with heat-based eviction and a persona module, thereby ensuring that topic-aligned content remains accessible while maintaining consistency with users' specific preferences.

(3) Our proposed MemoryOS achieves superior performance across all benchmark datasets due to its hierarchical storage design, semantic retrieval capabilities, and persona-driven dynamic updating, which ensure coherent and accurate memory management. 
Notably, the model’s advantages are particularly pronounced in more challenging memory management tasks. For example, on the LoCoMo benchmark with gpt-4o-mini, it achieves average improvements of 49.11\%  on  F1 score and 46.18\%  on BLEU-1,  while on the easier GVD dataset, in which all methods achieve higher baseline accuracy, our MemoryOS still surpasses the SOTA baseline A-Mem by 3.2\%  in accuracy, showing robust handling of complex long-context tasks requiring semantic consistency.

(4) 
To evaluate model efficiency, we employed two metrics: tokens consumed ( in memory retrieval) and average LLM calls in each response. As shown in Table \ref{tab:consumption_analysis}, our method outperforms the Top-2 baselines (i.e., MemGPT and A-Mem) in both aspects, requiring significantly fewer LLM calls than A-Mem* (4.9 vs.13) and much lower token consumption than MemGPT (3,874 vs. 16,977). 

\begin{figure}
\centering
\includegraphics[width=\linewidth]{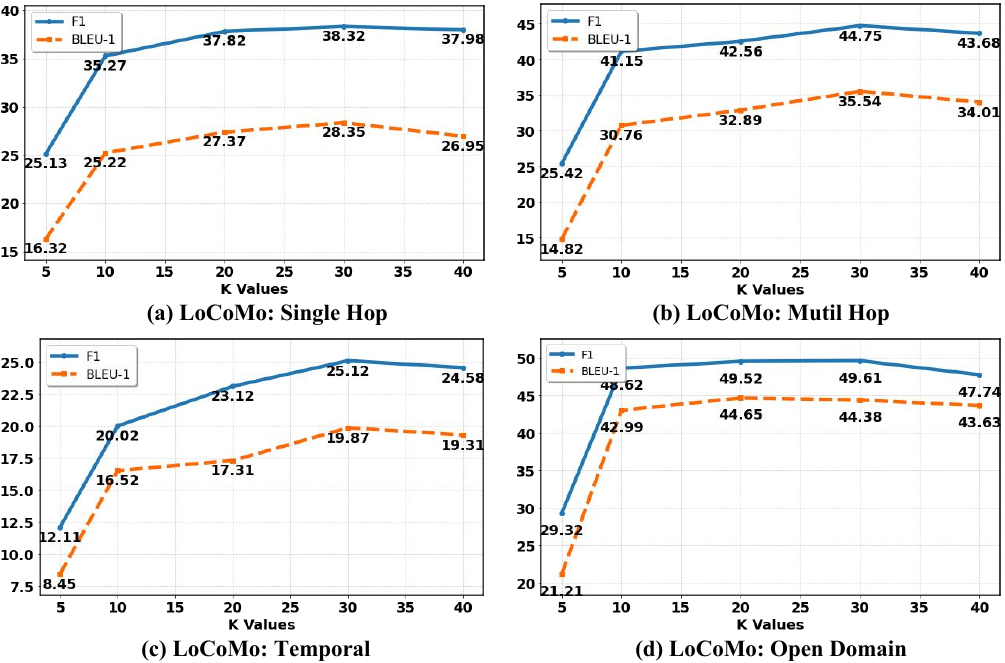}
\caption{Impact of hyperparameter $k$ (retrieved pages in MTM) on LoCoMo benchmark.}
\label{fig:mem_topk}
\end{figure}

\begin{figure*}
\centering
\includegraphics[width=\linewidth]{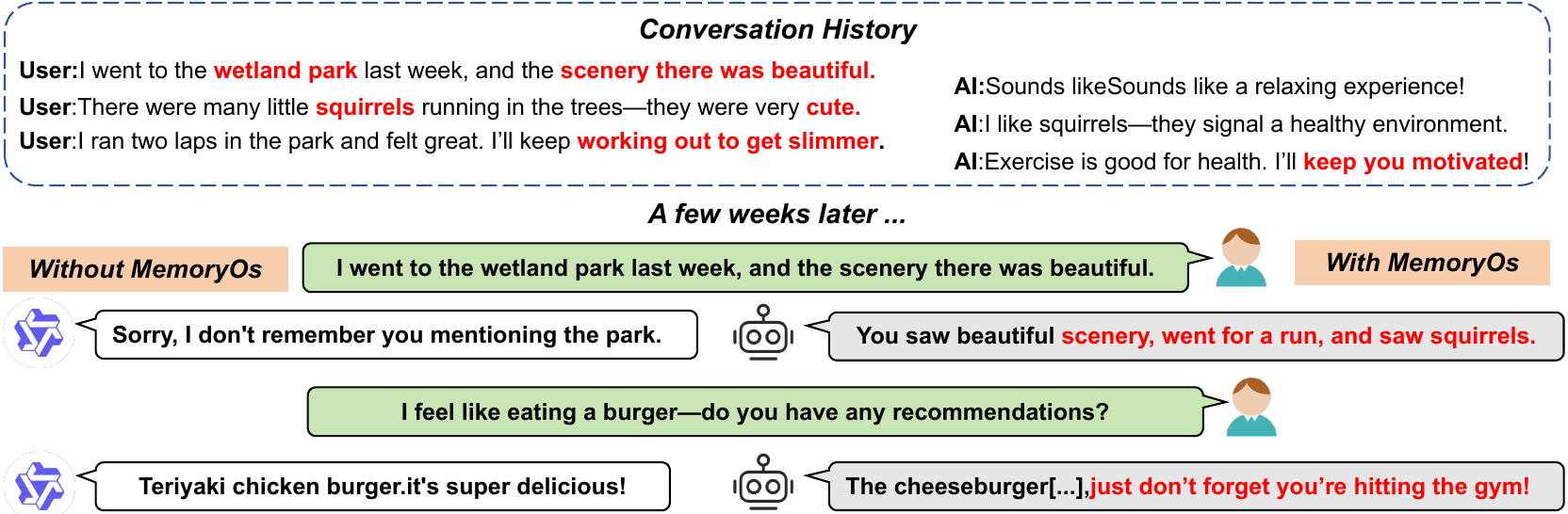}
\caption{Case study demonstrating the positive impact of introducing our memory management system. Left: default LLMs; Right: with MemoryOS.}
\label{fig:mem_case}
\end{figure*}

\subsection{Ablation Study}

To assess the contribution of each core module in our framework, we perform an ablation study by individually removing three key components: the Mid-Term Memory (-MTM), the Long-term Persona Module (-LPM), and the Diague page Chain (-Chain) and the entire memory system (-MemoryOS).  The results are presented in Figure~\ref{fig:ablation_gvd}. 
We can see that the memory system plays a pivotal role in the quality of responses during long dialogues. Without MemoryOS, the model's performance drastically reduces. In MemoryOS, the Mid-Term Memory (MTM) has the most significant impact, followed by the Long-Term Memory (LPM), while the Chain has the least impact. 



\subsection{Hyperparameter Analysis}
We analyze the impact of top-k retrieved dialogue pages from Mid-Term Memory (MTM) on model performance.
As shown in Fig.~\ref{fig:mem_topk}, by setting the hyperparameter $k$ with different values $k=\{5,10,20,30,40\}$ on the LoCoMo benchmark, we can see that
the model's performance improves as $k$ increases, but the improvements diminish when exceeding a threshold. Retrieving more pages can enhance model performance, but excessive content may introduce noise, adversely affecting performance. We set $k=10$ to achieve a relatively favorable performance while minimizing computational overhead.





\subsection{Case Study}
To visually demonstrate the role of our memory system, specifically how user long-term memory maintenance enhances conversational consistency. We present case studies in Fig.~\ref{fig:mem_case}. Based on the conversation history, we show the responses of default LLMs and our LLMs with MemoryOS. 
We can see that MemoryOS exhibits excellent capabilities in recalling users' long-term conversations and preferences.
For example, MemoryOS recalls details like "seeing the scenery, running, and spotting squirrels in the wetland park."  from the initially mentioned a few weeks ago,  “I went to the wetland park ...". 
These details are retrieved through the interplay and mutual support of mid-term memory's segment-page storage and the dialogue page chain.
In addition, since the system integrates a personalization module, it can remember the user's goal of “wanting to get fit” and later proactively reminds the user when they express a desire to eat a burger: “Don’t forget you want to get slimmer”. 
This highlights the crucial role of the memory module in enhancing dialogue coherence and user experience.




\section{Conclusion}
Inspired by memory management mechanisms in operating systems, we pioneers propose a novel memory management system, MemoryOS, for AI agents. Implemented by a hierarchical memory storage architecture, MemoryOS addresses the fixed context window limitations in long conversations.
By adapting OS-style segment-paging storage for dialogue history, MemoryOS enables efficient memory storage, updating, and semantic retrieval using heat-driven eviction to dynamically prioritize critical information across memory tiers. 
The integrated persona module captures evolving user preferences via personalized trait extraction, ensuring responses align with long conversation contexts.
By bridging OS principles with AI memory management, MemoryOS empowers LLMs to sustain coherent, personalized conversations over extended interactions, enhancing human-like dialogue capabilities in real-world applications.

\bibliography{Reference}

\begin{thebibliography}{21}
\providecommand{\natexlab}[1]{#1}

\bibitem[{Bensoussan et~al.(1972)Bensoussan, Clingen, and Daley}]{bensoussan1972multics}
Andr{\'e} Bensoussan, C.~T. Clingen, and R.~C. Daley. 1972.
\newblock The multics virtual memory: Concepts and design.
\newblock \emph{Communications of the ACM}, 15(5):308--318.

\bibitem[{Chhikara et~al.(2025)Chhikara, Khant, Aryan, Singh, and Yadav}]{chhikara2025mem0}
Prateek Chhikara, Dev Khant, Saket Aryan, Taranjeet Singh, and Deshraj Yadav. 2025.
\newblock Mem0: Building production-ready ai agents with scalable long-term memory.
\newblock \emph{arXiv preprint arXiv:2504.19413}.

\bibitem[{DeepSeek-AI et~al.(2025)DeepSeek-AI, Guo, Yang, Zhang, Song, Zhang, Xu, Zhu, Ma, Wang, Bi, Zhang, Yu, Wu, Wu, Gou, Shao, Li, Gao, Liu, Xue, Wang, Wu, and Feng}]{deepseekai2025deepseekr1incentivizingreasoningcapability}
DeepSeek-AI, Daya Guo, Dejian Yang, Haowei Zhang, Junxiao Song, Ruoyu Zhang, Runxin Xu, Qihao Zhu, Shirong Ma, Peiyi Wang, Xiao Bi, Xiaokang Zhang, Xingkai Yu, Yu~Wu, Z.~F. Wu, Zhibin Gou, Zhihong Shao, Zhuoshu Li, Ziyi Gao, and 5 others. 2025.
\newblock \href {https://arxiv.org/abs/2501.12948} {Deepseek-r1: Incentivizing reasoning capability in llms via reinforcement learning}.
\newblock \emph{Preprint}, arXiv:2501.12948.

\bibitem[{Denning(1970)}]{denning1970virtual}
Peter~J. Denning. 1970.
\newblock Virtual memory.
\newblock \emph{ACM Computing Surveys}, 2(3):153--189.

\bibitem[{Du et~al.(2025)Du, Huang, Zheng, Wang, Montella, Lapata, Wong, and Pan}]{du2025rethinking}
Yiming Du, Wenyu Huang, Danna Zheng, Zhaowei Wang, Sebastien Montella, Mirella Lapata, Kam-Fai Wong, and Jeff~Z Pan. 2025.
\newblock Rethinking memory in ai: Taxonomy, operations, topics, and future directions.
\newblock \emph{arXiv preprint arXiv:2505.00675}.

\bibitem[{Huang et~al.(2024)Huang, Lan, Sun, Shi, and Bai}]{emotional}
Le~Huang, Hengzhi Lan, Zijun Sun, Chuan Shi, and Ting Bai. 2024.
\newblock \href {https://doi.org/10.1109/ICKG63256.2024.00023} {Emotional rag: Enhancing role-playing agents through emotional retrieval}.
\newblock In \emph{2024 IEEE International Conference on Knowledge Graph (ICKG)}, pages 120--127.

\bibitem[{Li et~al.(2024)Li, Yang, Zhang, Deng, Wang, and Chua}]{li2024hello}
Hao Li, Chenghao Yang, An~Zhang, Yang Deng, Xiang Wang, and Tat-Seng Chua. 2024.
\newblock Hello again! llm-powered personalized agent for long-term dialogue.
\newblock \emph{arXiv preprint arXiv:2406.05925}.

\bibitem[{Li et~al.(2025)Li, Guan, Wu, Wu, and Yan}]{mayipersona}
Jia-Nan Li, Jian Guan, Songhao Wu, Wei Wu, and Rui Yan. 2025.
\newblock \href {https://arxiv.org/abs/2503.15463} {From 1,000,000 users to every user: Scaling up personalized preference for user-level alignment}.
\newblock \emph{Preprint}, arXiv:2503.15463.

\bibitem[{Liu et~al.(2023)Liu, Yang, Shen, Hu, Zhang, Gu, and Zhang}]{liu2023think}
Lei Liu, Xiaoyan Yang, Yue Shen, Binbin Hu, Zhiqiang Zhang, Jinjie Gu, and Guannan Zhang. 2023.
\newblock Think-in-memory: Recalling and post-thinking enable llms with long-term memory.
\newblock \emph{arXiv preprint arXiv:2311.08719}.

\bibitem[{Maharana et~al.(2024)Maharana, Lee, Tulyakov, Bansal, Barbieri, and Fang}]{maharana-etal-2024-evaluating}
Adyasha Maharana, Dong-Ho Lee, Sergey Tulyakov, Mohit Bansal, Francesco Barbieri, and Yuwei Fang. 2024.
\newblock \href {https://doi.org/10.18653/v1/2024.acl-long.747} {Evaluating very long-term conversational memory of {LLM} agents}.
\newblock In \emph{Proceedings of the 62nd Annual Meeting of the Association for Computational Linguistics (Volume 1: Long Papers)}, pages 13851--13870, Bangkok, Thailand. Association for Computational Linguistics.

\bibitem[{Ocker et~al.(2025)Ocker, Deigmöller, Smirnov, and Eggert}]{ocker2025ground}
Felix Ocker, Jörg Deigmöller, Pavel Smirnov, and Julian Eggert. 2025.
\newblock \href {https://arxiv.org/abs/2505.06328} {A grounded memory system for smart personal assistants}.
\newblock \emph{Preprint}, arXiv:2505.06328.

\bibitem[{Packer et~al.(2023)Packer, Fang, Patil, Lin, Wooders, and Gonzalez}]{packer2023memgpt}
Charles Packer, Vivian Fang, Shishir\_G Patil, Kevin Lin, Sarah Wooders, and Joseph\_E Gonzalez. 2023.
\newblock Memgpt: Towards llms as operating systems.

\bibitem[{Papineni et~al.(2002)Papineni, Roukos, Ward, and Zhu}]{papineni2002bleu}
Kishore Papineni, Salim Roukos, Todd Ward, and Wei-Jing Zhu. 2002.
\newblock Bleu: a method for automatic evaluation of machine translation.
\newblock In \emph{Proceedings of the 40th annual meeting of the Association for Computational Linguistics}, pages 311--318.

\bibitem[{Park et~al.(2023)Park, O'Brien, Cai, Morris, Liang, and Bernstein}]{aitown2024}
Joon~Sung Park, Joseph O'Brien, Carrie~Jun Cai, Meredith~Ringel Morris, Percy Liang, and Michael~S Bernstein. 2023.
\newblock Generative agents: Interactive simulacra of human behavior.
\newblock In \emph{Proceedings of the 36th annual acm symposium on user interface software and technology}, pages 1--22.

\bibitem[{Wang et~al.(2025)Wang, Liang, Yang, Huang, Wu, Wu, Lu, Ma, and Li}]{wang2025scmenhancinglargelanguage}
Bing Wang, Xinnian Liang, Jian Yang, Hui Huang, Shuangzhi Wu, Peihao Wu, Lu~Lu, Zejun Ma, and Zhoujun Li. 2025.
\newblock \href {https://arxiv.org/abs/2304.13343} {Scm: Enhancing large language model with self-controlled memory framework}.
\newblock \emph{Preprint}, arXiv:2304.13343.

\bibitem[{Wu et~al.(2025)Wu, Liang, Zhang, Wang, Zhang, Guo, Tang, and Liu}]{wu2025human}
Yaxiong Wu, Sheng Liang, Chen Zhang, Yichao Wang, Yongyue Zhang, Huifeng Guo, Ruiming Tang, and Yong Liu. 2025.
\newblock From human memory to ai memory: A survey on memory mechanisms in the era of llms.
\newblock \emph{arXiv preprint arXiv:2504.15965}.

\bibitem[{Xu et~al.(2025)Xu, Liang, Mei, Gao, Tan, and Zhang}]{xu2025mem}
Wujiang Xu, Zujie Liang, Kai Mei, Hang Gao, Juntao Tan, and Yongfeng Zhang. 2025.
\newblock A-mem: Agentic memory for llm agents.
\newblock \emph{arXiv preprint arXiv:2502.12110}.

\bibitem[{Yuan et~al.(2024)Yuan, Wang, Feng, Pan, Li, Wang, Miao, and Li}]{yuan-etal-2024-generative}
Peiwen Yuan, Xinglin Wang, Shaoxiong Feng, Boyuan Pan, Yiwei Li, Heda Wang, Xupeng Miao, and Kan Li. 2024.
\newblock Generative dense retrieval: Memory can be a burden.
\newblock In \emph{Proceedings of the 18th Conference of the European Chapter of the Association for Computational Linguistics (Volume 1: Long Papers)}, St. Julian{'}s, Malta. Association for Computational Linguistics.

\bibitem[{Zhang et~al.(2024)Zhang, Bo, Ma, Li, Chen, Dai, Zhu, Dong, and Wen}]{zhang2024survey}
Zeyu Zhang, Xiaohe Bo, Chen Ma, Rui Li, Xu~Chen, Quanyu Dai, Jieming Zhu, Zhenhua Dong, and Ji-Rong Wen. 2024.
\newblock A survey on the memory mechanism of large language model-based agents.
\newblock \emph{arXiv preprint arXiv:2404.13501}.

\bibitem[{Zheng et~al.(2020)Zheng, Zou, and Wang}]{zheng2020segmentpage}
Yan Zheng, Tong Zou, and Xingyan Wang. 2020.
\newblock Segment-page-combined memory management technology based on a homegrown many-core processor.
\newblock \emph{CCF Transactions on High Performance Computing}, 2(4):376--381.

\bibitem[{Zhong et~al.(2024)Zhong, Guo, Gao, Ye, and Wang}]{zhong2024memorybank}
Wanjun Zhong, Lianghong Guo, Qiqi Gao, He~Ye, and Yanlin Wang. 2024.
\newblock Memorybank: Enhancing large language models with long-term memory.
\newblock In \emph{Proceedings of the AAAI Conference on Artificial Intelligence}, volume~38, pages 19724--19731.

\end{thebibliography}

\end{document}